\documentclass[conference]{IEEEtran}
\IEEEoverridecommandlockouts
\usepackage{cite}
\usepackage{amsmath,amssymb,amsfonts}
\usepackage{algorithmic}
\usepackage{graphicx}
\usepackage[table,xcdraw]{xcolor}
\usepackage{textcomp}
\usepackage{tabularx}
\usepackage{xcolor}
\usepackage{caption}
\def\BibTeX{{\rm B\kern-.05em{\sc i\kern-.025em b}\kern-.08em
    T\kern-.1667em\lower.7ex\hbox{E}\kern-.125emX}}
\begin{document}
\title{ NUBOT: Embedded Knowledge Graph With RASA Framework for Generating Semantic Intents Responses in Roman Urdu 
\\
}
\author{\IEEEauthorblockN{\textsuperscript{} Johar Shabbir}
\IEEEauthorblockA{\textit{Computer Science.FAST University} \\
Islamabad, Pakistan \\
i192070@nu.edu.pk}
\and
\IEEEauthorblockN{\textsuperscript{} Muhammad Umair Arshad}
\IEEEauthorblockA{\textit{Computer Science.FAST University} \\
Islamabad, Pakistan \\
umair.arshad@nu.edu.pk}
\and
\IEEEauthorblockN{\textsuperscript{} Waseem Shahzad}
\IEEEauthorblockA{\textit{Computer Science.FAST University} \\
Islamabad, Pakistan \\
waseem.shahzad@nu.edu.pk}
}
\maketitle
\begin{abstract}
The Internet is historically modeled as a graph, where each collection of protocol layers are implemented by the node and each edge corresponds to relation to physical contact. Regrettably, as compared This model, with the real Internet, falls short. In the traditional world of the Model, one or more static IP addresses address the nodes.
Understanding the topology of the Internet in terms of its connectivity in the face of these major extensions to the classical model The mission has become a challenging one. It has become difficult to identify basic concepts, such as relationships between neighbors and peers. The more complex forwarding and routing processes are the only ones.
In the past, researchers have also attempted to generalize Internet concepts, recognizing the failure of the classical model to adequately describe ground realities and conceptual isomorphism.
The understanding of the human language is quantified by identifying intents and entities. Even though classification methods that rely on labeled information are often used for the comprehension of language understanding, it is incredibly time consuming and tedious process to generate high propensity supervised datasets. In this paper, we present the generation of accurate intents for the corresponding Roman Urdu unstructured data and integrate this corpus in RASA NLU module for intent classification. We embed knowledge graph with RASA Framework to maintain the dialog history for semantic based natural language mechanism for chatbot communication. We compare results of our work with existing linguistic systems combined with semantic technologies. Minimum accuracy of intents generation is 64 percent of confidence and in the response generation part minimum accuracy is 82.1 percent and maximum accuracy gain is 96.7 percent. All the scores refers to log precision, recall, and f1 measure for each intents once summarized for all. Furthermore, it creates a confusion matrix represents that which intents are ambiguously recognized by approach. 
\end{abstract}
\section{Introduction}
One of the reason that organizations fail to achieve effectivity and customer satisfaction is information inadequacy \cite{duong2017natural}\cite{javed2020collaborative}. Information is one of the most essential resources in modern times, as it guides human thinking, planning and subsequent actions. 
Major causes of information inadequacy is that information exist but cannot be found or delivered, or information transmission is delayed. To handle these problems, chatbot is one of the best solutions which can help both, organizations and clients to achieve their goals more effectively\cite{beg2013constraint}\cite{beg2001memory}\cite{khawaja2018domain}. A chatbot can automate customer support for similar queries, save human resources for qualitative tasks, accelerate operations, give better user interaction, and are easy to use, cost effective, and time efficient.
Intent based systems are those system that are based on natural language of human, These systems identify and predict the intention of the user and predicate that what user actually want to say in specific context. RASA Framework has two major components named as RASA NLU that perform intent classification and entity extraction from the training dataset and second component is Dialog Management Model (DMM) that generate the specific response according to intents that are classify also entities that are extracted\cite{beg2008critical}.
We embedded Knowledge graph with RASA Dialog Management Model (DMM), Knowledge Graph also helps the DMM to generate the response by following the dialog history of the user and generate the response according to user intention.
Memory is a very important factor in conversations. The context from previous dialogues emerges with the new concepts to develop a new context\cite{khawaja2018domain}. Simple deep learning models are useful to answer questions but to carry out information from previous dialogues to form new dialogue is not feasible\cite{asad2020deepdetect}\cite{zafar2019using}\cite{naeem2020deep}. To solve this problem RNN \cite{bottou2014machine} and LSTM \cite{sundermeyer2012lstm}\cite{gunasekara2019uniontbot} were introduced.
In the 1960s the first chatbot was created at MIT named ELIZA \cite{weizenbaum1966eliza} which worked on the basic principle of identifying keywords and generating an output closely related to those keywords. Since ELIZA a lot of new frameworks and algorithms have been introduced for rule mining from the data and generating a response based on those rules. The rule mining approach\cite{hussain2019survey} is good for small applications with a limited amount of data\cite{arshad2019corpus} but today it is normal for an application to process from complex and big data \cite{naeem2020deep}\cite{asad2020deepdetect}\cite{hatcher2018survey}. Using this technique, words are clustered together and new incoming words are classified accordingly which helps in generating a response for the user. Words embeddings are also used to check sentence structures, grammatical errors, translations \cite{zou2013bilingual}, etc.
The proposed approach aims to develop an Roman Urdu enquires chatbot with embedded knowledge graph in RASA Framework, also we tried to automate the process of intent generation by using AI and Deep Learning techniques so that it will reduces the human effort for training dataset. Our results shows the superiority over the existing approaches.

In comparison to the organizations, higher management schemes invest immense amount of money to retain their exposure in the targeted markets. Stakeholders will pose massive amounts of queries from the web media.In many cases, several requests are neglected or have late response due to lack of human resources or variation in the time zone.So,we proposed semantic base mechanism with natural language processing to generate semantically correct generation of the response for the chatbot systems for Roman Urdu Data-set.


\begin{figure}[h!]
\centering
  \includegraphics[width=0.4\textwidth]{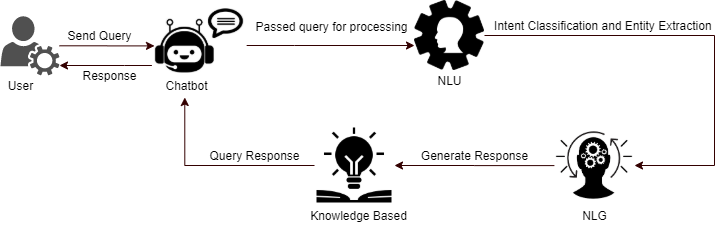}~\\[0.3cm]
  \caption{Represents the proposed high level diagram of end to end communication of chatbot in RASA Framework embedded knowledge graph that helps out in the process of accurate response generation.}
\end{figure}
\section{Related Work}
Existing systems are discussed in the section underneath as well as the systems are divided into three main categories namely, OWL verbalizer libraries, OWL verbalizer projects, and Ontology-based chatbots\newline
Social media, today, shows the exponential development of modern
Society as it becomes the predominant medium to connect and express itself for Internet users \cite{loo2005declarative}\cite{beg2010graph} \cite{zafar2020search} \cite{dilawar2018understanding}. Around the world, people use to access the Internet, set up a range of devices and services, Social networks, online business results, e-commerce, e-surveys, etc. Social networking is currently not only a technology that offers data to customers\cite{zafar2019using}\cite{zafar2018deceptive}\cite{javed2020alphalogger}.
\subsection{API based NLG systems}
\subsubsection{SimpleNLG:}
SimpleNLG is a library with Java API that generates sentences after the subject-verb and the subjects are described. The Library follows the API approach to sentence making and words must be transmitted into the Java System. Knowledge of programming is required for setting inputs to a Java program library. This is a good approach for a  small domain approach since the lexicons will be specified by the user, even though the user is expected to have a level design. However, for the Ontologies written in the OWL file, the article claims not specifically tested on the Ontology of OWL. Hence for our use which case generates a chatbots API using Ontologies.
\subsection{Ontology based chatbots}
\subsubsection{ Ontology to Relational Database Mapping:}
To store ontology data and conduct queries on that several solutions exist in libraries, such as storing they are related, entity, or entity-linked. space. Ontologies are less simple in relational databases than on Place ontologies of libraries related to events or artifacts, Since relational database management systems are not usable support inheritance. However, relational database management systems have significant advantages over object or object relational database management systems. In particular,relational database management systems provide maturity,performance, robustness, reliability, and availability, and that’s what pushed us to go with the relational database option.Several studies \cite{astrova2007storing}\cite{vysniauskas2006transforming} have been conducted regarding the mapping between ontologies and relational databases. It is out of our scope to go in details with each study since the
proposed approach will work over the result of this mapping and go further in its process\cite{zhuge2008resource}.
\subsubsection{OntBot: Ontology based Chatbot:}
A chatbot is a computer program which interacts with
users using natural language generated from a machine perspective. Ontologies enable cooperation between the human and the computer environment \cite{al2011ontbot}. While ontologies are helpful for the chatbot's database portion, which reduces computational costs and searches data seamlessly, the natural language generation part consists of many advanced techniques that will also have a higher computational cost.
\subsection{Natural Language Generation}
The natural language generation is the means of generating effective texts in a machine perspective for human readability \cite{di2012inlg}. The major work was carried out for the Natural Language Generation from the 1970's onward. Goldman works on the lexicalizing an underlying conceptual materials. Another work carried out is providing a description for a tic-tac-tie game by Davey in 1979. These are considered as some of the first contributions focused on NLG \cite{dale1998introduction}. Generation of response output can be pre-structured or in a completely unstructured form.
\subsubsection{Static responses:}
This is the simplest way of generating the text for the NLG component. The generated response is already predefined by the system, where the variables of the sentences are replaceable. This type of response could be used as a template.
\subsubsection{Dynamic responses:}
This is another approach to generate text is by using resources. This can be a knowledge base component with a decision-making component that specifies the score for the nearest generated response by matching up with the user query. This is mostly used with Q and A systems \cite{toniuc2017climebot}.
\subsubsection{Generated responses:}
By using this approach a huge amount of text conversational examples is taken into a deep learning technique to train the machine to generate responses \cite{aliane2010khalil}. huge data would be needed in order to provide more accurate responses, and it would also incur a higher computational cost. Sometimes the generated responses would be irrelevant, yet with more training and defining, more rules would give more accurate information \cite{couto2017building}. \newline
Even though there exist many representations over the years for linguistic surface-oriented definition over semantic representations, the natural representation of NLG is considered as semantic representation. Because of the characteristics of the semantic web machine-processable paradigm, it has been an interesting topic to NLG enthusiasts \cite{staykova2014natural}.
\section{Methodology}
Building a contextual, interactive, and intelligent chatbot handle dynamic queries and generate dynamic responses is not an easy task. It requires computer to understand human language. Computer must understand human language’s semantics, syntax, and pragmatics etc. Moreover, everyone has a different way of typing a message and not all the users can follow same format for writing queries.\\
A chatbot talking to a human being, which is a completely unique individual, raises problems like handling different usage of slangs, habit of misspelling certain words, usage of short forms and cool words etc. Natural Language Processing is not yet developed enough to be able to handle conversation in resource-poor local languages like Roman Urdu. Users are not consistent with their language and their goals. It is very difficult to predict everything a user can ask and every way a user can ask a question in. So, handling randomness of a human being is also a major problem for a chatbot developer. A user expects a chatbot to be as close to human intelligence as possible for a better experience. The most difficult task for a chatbot is to understand the intent of the user (what the user is looking for) and remember the context of the whole conversation.\\
All the limitations discussed above make it difficult to develop a chatbot that can facilitate its users effectively and efficiently specially in resource poor languages like Roman Urdu.
For the purpose of domain knowledge, we use Literature Surveys to understand several categories i.e. Natural language generation, Semantic-based natural language generation, knowledge graph-based response generation.\newline
\begin{figure}[h!]
\centering
  \includegraphics[width=0.4\textwidth]{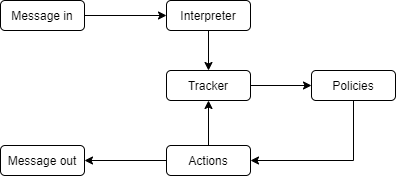}~\\[0.3cm]
  \caption{Represents the workflow of generated response in core RASA Framework. It represent the complete process of end to end conversation between user and the chatbot.}
\end{figure}
Knowledge extract from the research we construct a theory that "If we are able to generate semantically correct language generation than it cause lower computational cost with an intelligent approach for the system". For the initialization of the project, we used the PRINCE2 project management methodology because it is flexible with our research project.\\
\begin{figure}[h!]
\centering
  \includegraphics[width=0.45\textwidth]{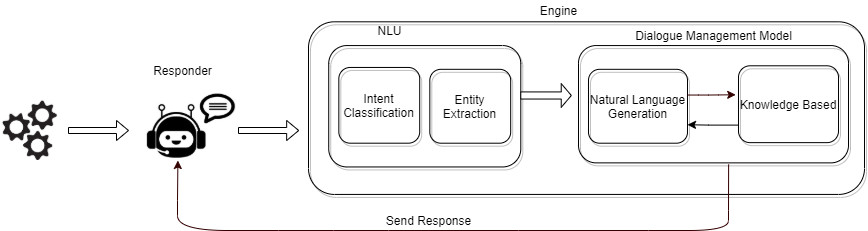}~\\[0.3cm]
  \caption{Represents the proposed Architecture Diagram with RASA NLU which performs two basic tasks of intent classification and entity extraction also Dialog Management Model embedded with knowledge graph for generating response }
\end{figure}
We are using the RASA framework for developing a chatbot. RASA is a state-of-the-art and open-source framework platform for the development of chatbot. We performed intent generation using Latent Dirichlet Allocation (LDA) as topic modeling for the intent labeling. In the NLU component, We have used Whitespce Tokkenizer the tokenizer breaks text into terms whenever it encounters a whitespace character as a separator, Count Vector Featurizer at the character level that Creates bag-of-words representation of user messages, intents and responses, and  Dual Intent Entity Transformer (DIET) used for intent classification and entity extraction. We used REGEX Featurizer, DIET Classifier, Entity Synonym Mapper, and Lexical Syntactic Featurizer for the entity extraction. In the NLG component, we used three types of RASA Policies namely Memoization policy, Fallback policy and TED policy for the purpose of generating the response. RASA Provide an environment where we used multiple policies in a single configuration and customize our policies according to end goal. Mainly we embedded knowledge graph with RASA Framework to Make it more efficient for our domain. A knowledge graph break the query into tokens and helps out to generating contextual information helpful in the process of response generation.
\begin{figure}[h!]
\centering
  \includegraphics[width=0.5\textwidth]{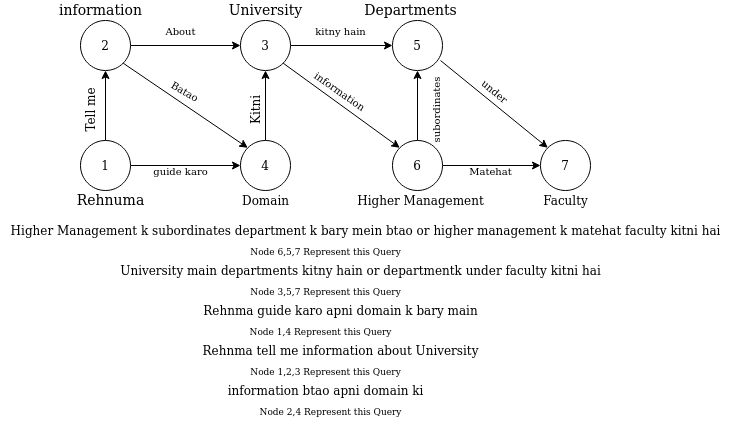}~\\[0.4cm]
  \caption{Sample Representation of Roman Urdu Query in a knowledge Graph}
\end{figure}

First of all the output of the NLU component in the form of intent classification and entity extraction treated as the input of the NLG component. For the sake of generating the response, the input is passed through the Memoization policy if the corresponding intent existed than it will generate response regards to the corresponding query otherwise it will be passed to the Fallback policy if the corresponding response exists than it will generate a response in some case it will not generate a response than input is passed to TED policy and if it fails to identify the response than finally response is generated from the knowledge graph.

\begin{figure}[h!]
\centering
  \includegraphics[width=0.45\textwidth]{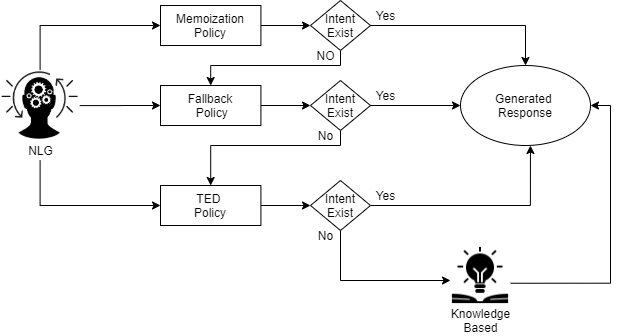}~\\[0.3cm]
  \caption{Represents the architecture of the NLG components for generating response for the chatbot }
\end{figure}
Even though the traditional NLG needs to be trained with higher computational cost, the component needs to have more data and more conversations in order to provide semantically correct response. But if the chatbot can understand the context of the stored information, there can be a hypothesis that the computer would perform better with less computational cost and less filtering algorithms. For this reason we embedded knowledge graph with the RASA DMM component to maintain the context of conversation. Our results mirror the hypothesis for less computation.
\section{Evaluation And Results}
\subsubsection{Dataset Shape:}
The dataset contains a nlu.md file that contains the intents for the NLU component.
Stories.md file contains stories RASA stories are the form of training data used to train the dialogue management model. A story represents the conversation between a user and an AI assistant. A training example for the RASA Core dialogue system is called a story.\\
conversationtest.md file that contains specific tags this file contains tests to evaluate that your bot behavior as expected.config.yml file contains the configuration for NLU component.credentials.yml This file contains the credentials for the voice \& chat platforms in RASA framework and Responses.json This file contains the corresponding responses.
\begin{figure}[h!]
\centering
  \includegraphics[width=0.45\textwidth]{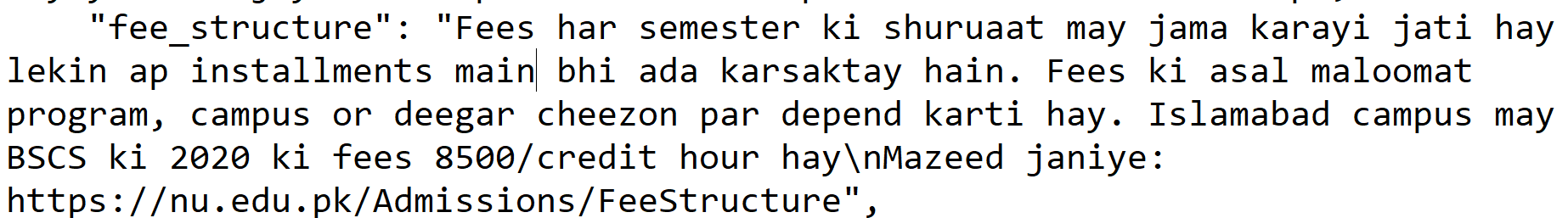}~\\[0.2cm]
  \caption{Represents the response related to the domain of the chatbot }
\end{figure}
\section{Results and Discussion}
\begin{figure}[h!]
\centering
  \includegraphics[width=0.25\textwidth]{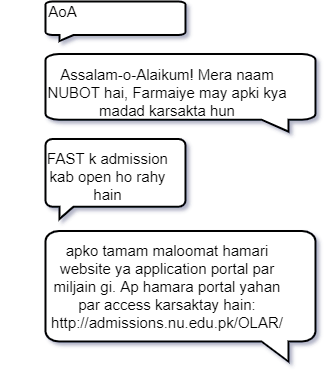}~\\[0.2cm]
  \caption{Represent the conversation detail between user and expert }
\end{figure}
\begin{figure}[h!]
\centering
  \includegraphics[width=0.25\textwidth]{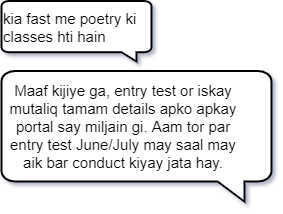}~\\[0.4cm]
  \caption{Represents output of the irrelevant query asked the expert }
\end{figure}
In this section, we have discussed that our focus is on developing a chatbot system for specific domain with minimum computational cost and produce accurate results. For this purpose, we evaluate our proposed work against intent generation confidence, F1 score and response generation F1 measure. we have constructed the confusion matrix for intents prediction against their features. the precision score for the confusion matrix is 85\%, the F1 score for the confusion matrix is 96.7\% and the F1 score is 89.1\% also from the Intent Prediction Confidence Distribution we got the confidence level of 0.96 for intent prediction against 130 unique samples of the intents for the purpose of generating response.
\begin{figure}[h!]
\centering
  \includegraphics[width=0.25\textwidth]{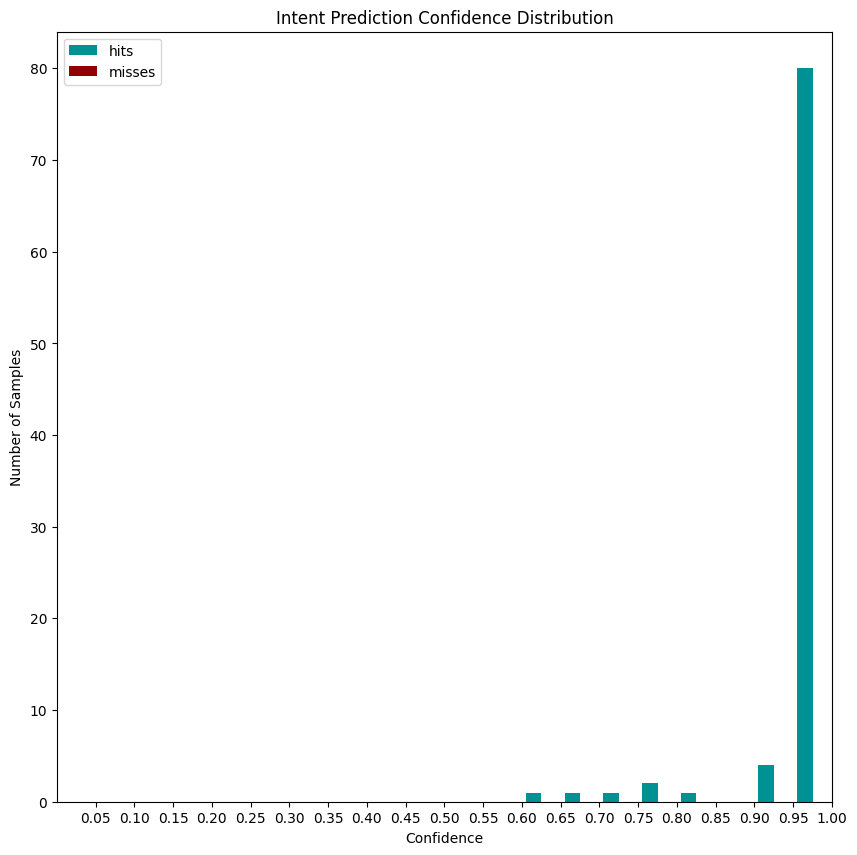}~\\[0.2cm]
  \caption{Represent the Intent Prediction Confidence Distribution using LDA with confidence level of 96.7\% F1 score }
\end{figure}

\begin{figure}[h!]
\centering
  \includegraphics[width=0.45\textwidth]{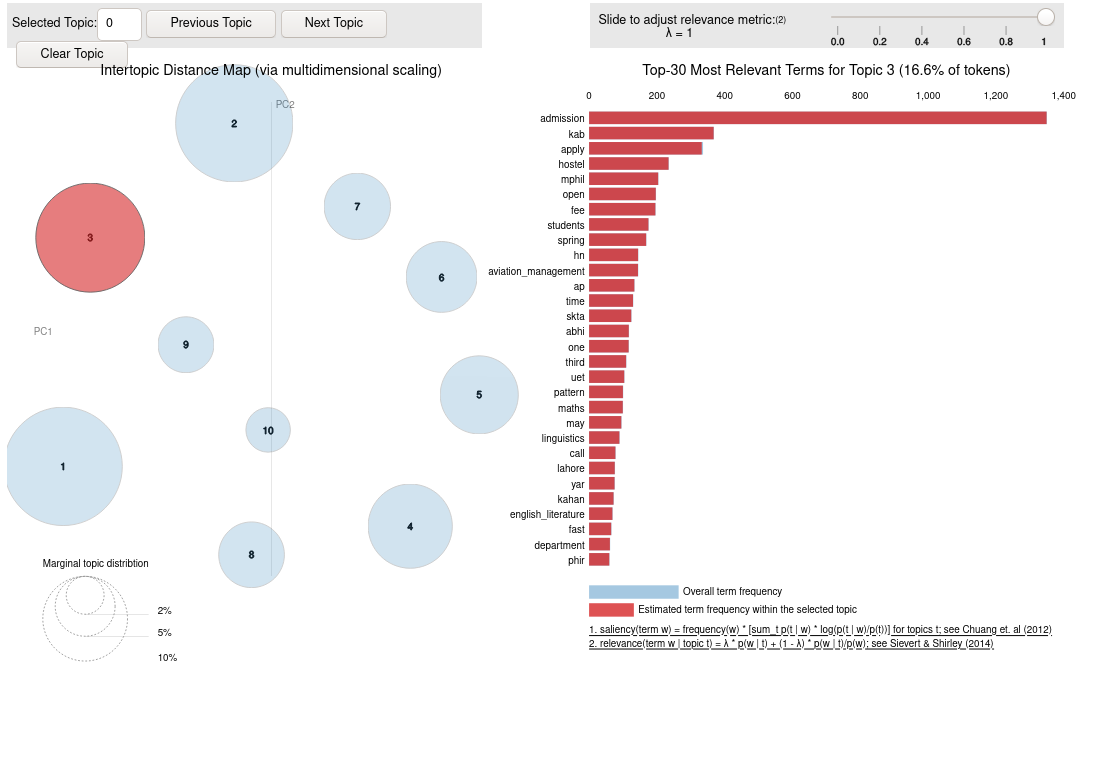}~\\[0.4cm]
  \caption{Represents the word distributions for a topic for different values of cluster size 10 for LDA with 16.7 percent of tokens }
\end{figure}

\begin{table}[h]
\centering
\resizebox{0.5\textwidth}{!}{%
\begin{tabular}{|
>{\columncolor[HTML]{EFEFEF}}l |
>{\columncolor[HTML]{EFEFEF}}l |
>{\columncolor[HTML]{EFEFEF}}l |}
\hline
\textbf{Evaluation Matrics} & \textbf{Approaches}               & \textbf{Scores}                   \\ \hline
\textbf{BLUE Score}         & Seq2Seq                           & 44.68                             \\ \hline
\textbf{\begin{tabular}[c]{@{}l@{}}Intent Generation Confidence\\ F1 Score\\ Response Generation F1 Score\end{tabular}} &
  Core RASA Framework &
  \textbf{\begin{tabular}[c]{@{}l@{}}79\%\\ 82.5\%\\ Max 81.1\%, Min 78.4\%\end{tabular}} \\ \hline
\textbf{\begin{tabular}[c]{@{}l@{}}Precision\\ Recall\\ F1\end{tabular}} &
  TCNGD &
  \begin{tabular}[c]{@{}l@{}}0.390\\ 0.705\\ 0.502\end{tabular} \\ \hline
\textbf{\begin{tabular}[c]{@{}l@{}}Intent Generation Confidence\\ F1 Score\\ Response Generation F1 Score\end{tabular}} &
  RASA Framework with knowledge graph &
  \textbf{\begin{tabular}[c]{@{}l@{}}85\%\\ 96.7\%\\ Max 89.1\%, Min 79.2\%\end{tabular}} \\ \hline
\textbf{F1 Score} &
  \begin{tabular}[c]{@{}l@{}}NER System\\ \\ Classification Systems\end{tabular} &
  \textbf{\begin{tabular}[c]{@{}l@{}}82.33  for intent recognition.\\ 97.3 for context extraction.\end{tabular}} \\ \hline
\end{tabular}%
}
\caption{Table Represent the different scores for Chatbots in poor resource languages and we get 85\% confidence on intent generation and 96.7\% F1 score on each classifying intents that helped out in the process of response generation. Also, Maximum confidence in each response is generated by the chatbot is 89.1\% and the minimum generated confidence of 79.2\% F1 score.}
\label{RASA Score Comparison}
\end{table}
Let 'K' be a hyper-parameter determining the number of clusters that we wanted to distribute our questions in. We ran LDA for different values of K such as 10, 20, 30 \& 40. We noticed that as the value of K increases, the size of individual clusters starts curtailing and there is more overlap between them. If K was set to a reasonable value (in our case 10 or 20), we were able to yield such clusters that the inter-topic distance mapping and the size of individual clusters was substantially good. For any cluster, LDA was able to produce a list of top key-terms which can really help give a label to that cluster and hence get an intent. Do note that prior to running LDA, preprocessing steps such as removing the stop words as well as too long/short questions were performed. 
\section{Conclusion}
The work is carried out based on the semantic technologies for the generation of the responses, and mostly the text generation has been a static method in most of the chatbots communication. To make better use of semantic-based technology, our solution would be an approach for generating response in appropriate way. If there is some out of scope query asked by the user than our chatbot will be able to tackle offensive as well as out of scope queries. \newline
knowledge gained from research, we propose a semantic-based natural language mechanism that can be applied to chatbots to generate text response for the user. Since our approach is a semantic-based, it can be easily integrated with any system as the future work, and a framework can be developed to be used for any domain to plug and play with any chatbot or any other system. Furthermore if we are able to generate specific domain knowledge graph from API call than we are able to make our chatbot more dynamic and flexible with users but here is the major issue of Roman Urdu lack of dataset from micro-blogging sites
.
\bibliographystyle{unsrt}
\bibliography{ref}
\vspace{12pt}
\color{red}
\end{document}